\documentclass[twoside]{article}
\usepackage[accepted]{original}
\usepackage[round]{natbib}
\renewcommand{\bibname}{References}
\renewcommand{\bibsection}{\subsubsection*{\bibname}}

\usepackage{hyperref}
\usepackage{url}

\usepackage[utf8]{inputenc} 
\usepackage[T1]{fontenc}    
\usepackage{hyperref}       
\usepackage{booktabs}       
\usepackage{amsfonts}       
\usepackage{nicefrac}       
\usepackage{microtype}      

\usepackage{subfiles}
\usepackage{amsmath}
\usepackage{graphicx}
\usepackage{subcaption}
\usepackage{float}
\usepackage{appendix}
\usepackage{multirow}
\usepackage[ruled]{algorithm2e}
\usepackage{color}
\usepackage{threeparttable}

\usepackage{subcaption}

\usepackage{amsthm}
\newtheorem{theorem}{Theorem}

\newcommand{\ours}{\textsl{Generator Surgery}}
\newcommand{\layers}{blocks}
\newcommand{\layer}{block}

\begin{document}

\runningauthor{Generator Surgery for Compressed Sensing}

\twocolumn[
\aistatstitle{Generator Surgery for Compressed Sensing}
\aistatsauthor{Niklas Smedemark-Margulies\textsuperscript{*}\footnotemark[1]
\And
Jung Yeon Park\textsuperscript{*}\footnotemark[1]
\And
Max Daniels\footnotemark[1]
\AND
Rose Yu\footnotemark[2]
\And
Jan-Willem van de Meent\footnotemark[1]
\And
Paul Hand\footnotemark[1]
}
\aistatsaddress{\footnotemark[1]Northeastern University \And  \footnotemark[2]University of California San Diego}
]

\newcommand{\fix}{\marginpar{FIX}}
\newcommand{\new}{\marginpar{NEW}}

\begin{abstract}
Image recovery from compressive measurements requires a signal prior for the images being reconstructed. Recent work has explored the use of deep generative models with low latent dimension as signal priors for such problems. However, their recovery performance is limited by high representation error. We introduce a method for achieving low representation error using generators as signal priors. Using a pre-trained generator, we remove one or more initial blocks at test time and optimize over the new, higher-dimensional latent space to recover a target image. Experiments demonstrate significantly improved reconstruction quality for a variety of network architectures. This approach also works well for out-of-training-distribution images and is competitive with other state-of-the-art methods. Our experiments show that test-time architectural modifications can greatly improve the recovery quality of generator signal priors for compressed sensing.
\end{abstract}

\section{Introduction}
\label{sec:introduction}

In inverse imaging problems, an estimated image signal is recovered from undersampled, possibly noisy measurements.
This problem is underdetermined, meaning that it is necessary to impose additional constraints or structural assumptions in order to recover the signal.
These additional structural assumptions are referred to as a ``signal prior''.
In this context, the use of the word ``prior'' generally does not refer to a distribution in the Bayesian sense. Rather, it encompasses any inductive biases or modeling assumptions that bias recovery towards the natural signal class.
Classical methods leverage sparsity or compressibility in a known basis \citep{candes-romberg-tao, wavelet-sparsity-book}.
Recent developments in density estimation and deep generative modeling, such as generative adversarial networks (GANs, \citet{goodfellow_gan}) and variational autoencoders (VAEs, \citet{vae}), have given rise to approaches that use a generator network as a signal prior.
Such signal priors map a low dimensional latent space to a manifold that approximates a particular class of natural images.
In this work, we focus on improving these generator network-based signal priors, which we collectively refer to as \emph{generator signal priors}.

Generator signal priors can compare favorably to sparsity priors at low undersampling ratios \citep{bora}. 
However, the low latent dimension of these networks results in high \emph{representation error}, i.e. there is a large discrepancy between the original image and the best possible recovered image.
This phenomenon has been observed in multiple architectures, even when reconstructing images drawn from the training set \citep{bora, inns}.
One path towards reducing the representation error is to train deep generative models with a higher latent dimension.
Unfortunately, training such models is difficult and unstable. This has motivated a variety of computationally expensive and/or complicated approaches
such as optimizing both the latent code and generator weights \citep{iagan} or adding an image-specific encoder before the generator and training it together with the latent code \citep{lcm}.

In this work, we address the problem of high representation error in generator signal priors and substantially improve their performance in compressed sensing with a simple technique that we call \ours{}.
Instead of altering the model's latent dimension or architecture at training time, we first train a deep generative model with low latent dimension, then ``cut'' one or more early \layers{} from the network, and treat the intermediate activations as a new latent image representation.
By cutting early \layers{} at inversion time, we increase the dimension of the latent representation and we gain increased image recovery quality at the cost of losing the generative sampling capability .
We show that this model with higher latent dimension exhibits dramatically reduced representation error and improved image recovery performance. 

We apply \ours{} to diverse neural network architectures and show that it is competitive with other recent methods for compressed sensing. Our method can be applied to pre-trained models (requiring only simple ad-hoc code patching) and is often simpler and computationally cheaper than other methods.

We further show that \ours{} can even recover high-quality images that are out-of-training-distribution while generator signal priors without surgery cannot. Here, we train models on one dataset, perform surgery, and then recover images from a different dataset. This is important in real-world applications like MRI imaging as they often include target images with novel features. 

The contributions of this paper are as follows:

\begin{itemize}
    \item We show that removing one or more initial \layers{} of a generator substantially increases recovery quality when using such models as signal priors for compressed sensing.
    \item Our method outperforms classical sparsity-based methods and is competitive with state-of-the-art methods for a similar number of parameters at inversion time. Our method also generalizes well to out-of-training-distribution images.
    \item Our experiments show that test-time architectural modifications can improve the performance of a model trained on one task (image generation) when applied to a different task (image recovery).
\end{itemize}
\section{Method}
\label{sec:method}
\subsection{Background}
We describe how a generator can be used as a signal prior for the problem of noisy compressed sensing (CS). All norms $\| \cdot \|$ are $L_2$ unless otherwise stated. A generative model or generator $\mathcal{G}_0: \mathbb{R}^{k_0} \mapsto \mathbb{R}^n$ maps a latent code $z_0$ drawn from a prior distribution $p(z_0)$ to an approximate sample from the data distribution. Typically the latent code is much lower in dimension that the output space, i.e. $k_0 \ll n$.

The task of compressed sensing of images is defined as follows. For an unknown target image $x \in \mathbb{R}^n$, we are given a measurement matrix $A \in \mathbb{R}^{m \times n}, m \ll n$ and measurements $y = Ax + \eta$, where $\eta$ is noise. Our goal is to find $\hat{x} \in \mathbb{R}^{n}$ that minimizes $\| x - \hat{x}\|$. Note that the problem is underdetermined, since $m < n$. Using the generator $G_0$, we can estimate $\hat{x}$ by finding a latent code $\hat{z}$ that optimizes the loss function:
\begin{align}
    \label{eqn:noisy_cs_objective}
    \min_{\hat{z}} \| y - AG(\hat{z}) \|.
\end{align}
Any optimization procedure can be used to optimize Eqn.~\eqref{eqn:noisy_cs_objective}. While the loss function is non-convex, gradient descent has been shown to work well \citep{bora}. In this paper, we study the standard case of a Gaussian measurement matrix $ A_{ij} \sim \mathcal{N}(0, 1/m)$ and zero-mean Gaussian noise $\eta$. 

\subsection{Generator Surgery (GS)}

\begin{figure}[t]
\begin{center}
\includegraphics[width=\columnwidth]{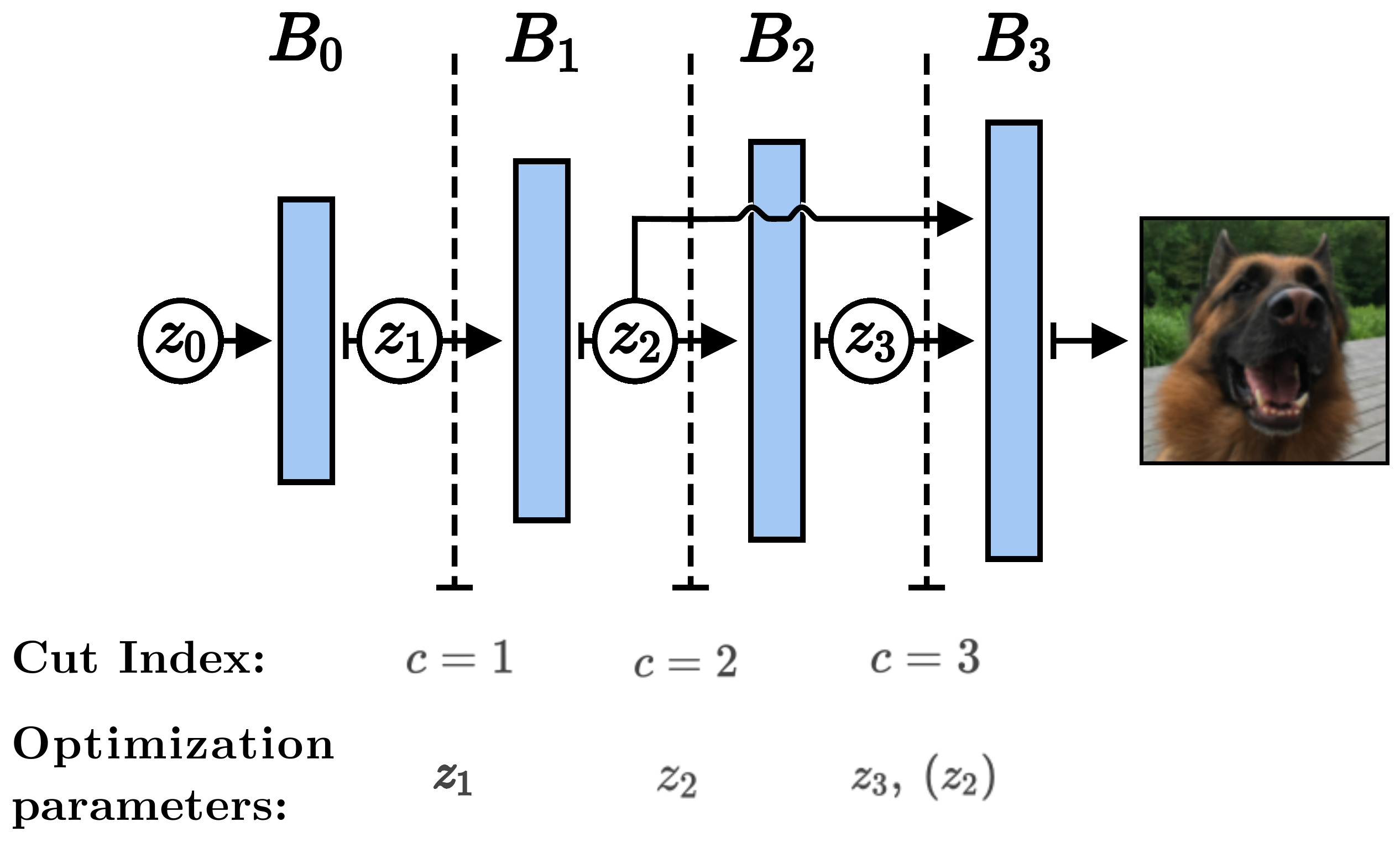}
\end{center}
\caption{\ours{}. To recover an image using a chosen cut index $c$, we remove earlier \layers{} $B_0 \ldots B_{c-1}$, and optimize all inputs to \layer{} $B_c$ (see Alg. \ref{alg:image-recovery}). This could include additional inputs such as skip connections, as shown for $c=3$.} 
\label{fig:method}
\vspace*{-\baselineskip}
\end{figure}

In our method, we assume that the generator $\mathcal{G}_0: \mathbb{R}^{k_0} \mapsto \mathbb{R}^n$ is composed of $d$ deterministic \layers{}:
\begin{align}
    \mathcal{G}_0(z_0) & = B_{d-1} \circ \ldots \circ B_0(z_0),
    \label{eqn:uncut-generator}
\end{align}
where we define $z_i \in \mathbb{R}^{k_i}$ as the input to \layer{} $B_i$. 
Each \layer{} $B_i$ is a differentiable function consisting of typical neural network operations (such as convolution, upsampling, and activation functions). We assume that the generator is expansive ($k_{i} \ge k_{0}, \forall i=1,\ldots d-1$), which holds for all architectures considered in this paper.
We refer to $\mathcal{G}_0$ as an ``uncut'' generator or ``No Generator Surgery (no GS)''.

After training the model in Eqn.~\eqref{eqn:uncut-generator} or using pre-trained weights, we cut away the first $c \ge 1$ \layers{} at test or inversion time, resulting in a network $\mathcal{G}_c: \mathbb{R}^{k_c} \mapsto \mathbb{R}^n = B_{d-1} \circ \ldots \circ B_{c}$ which accepts an intermediate representation $z_c$ as input. We refer to the resulting model as ``Generator Surgery (GS)'' or a ``cut'' generator. Since the generator is expansive, \ours{} increases the input dimension of the network and yields a more expressive model. Applying this method to compressed sensing and using gradient descent to optimize Eqn.~\eqref{eqn:noisy_cs_objective} results in Alg.~\ref{alg:image-recovery}.

It is important to point out that the cut generator is not a generative model (see Appendix). Sampling from the cut generator would require a prior $p(z_c)$ that matches the implicit pushforward measure described by $B_{c-1} \circ \dots \circ B_{0}(z_0)$ where $z_0 \sim p(z_0)$. In effect we are intentionally trading our ability to approximately sample from the learned distribution over images for expressivity of the latent representation in inversion tasks. Furthermore, as we are cutting layers from the original uncut generator, there are less backpropagation steps and thus computation is reduced compared to using the uncut generator.

\begin{algorithm}[!t]
\SetAlgoLined
\SetKwInOut{KwData}{Input}
\KwData{Pre-trained $\mathcal{G}_0$, cut index $c$, $A$, $y$, $\alpha$, restarts $R$, steps $T$}
\KwOut{estimated image $\hat{x}$}
$\mathcal{G}_c \gets cut(\mathcal{G}_0, c)$ 
\tcp*[r]{cut first $c$ \layers{}}
$L_{min} \gets \infty, \ \hat{x} \gets \vec{0}$ \;
\For{$r=1$ \KwTo $R$}{
    Initialize $z^{(0)}$ \;
    \For{$t=0$ \KwTo $T-1$}{
        $x^{(t)} \gets \mathcal{G}_c(z^{(t)})$ \tcp*[r]{estimate image}
        $L \gets \|y - Ax^{(t)}\|$ \tcp*[r]{compute loss}
        $z^{(t+1)} \gets z^{(t)} - \alpha \nabla_{z^{(t)}} L$ \tcp*[r]{update}
    }
    \tcp{keep result with min loss}
    \If{$L < L_{min}$}{
        $\hat{x} \gets x^{(T)}$ \\
        $L_{min} \gets L$
    }
}
\caption{Compressed Sensing using \ours}
\label{alg:image-recovery}
\end{algorithm}

\paragraph{Additional inputs.}
Our method is applicable to many different architectures, including those with skip connections or class-conditional models.
Specifically, if we cut the first $c$ \layers{} of a model and the new initial layer $B_{c}$ requires multiple inputs, all of these inputs are treated as optimizable parameters at inversion time. We use several architectures in experiments, one of which includes skip connections (BEGAN, \cite{began}) and another that includes class embeddings (BigGAN, \cite{biggan}).


\subsection{Theoretical Motivation}

For a generator $\mathcal{G}_0$ and a target image $x$, the \emph{representation error} is defined as
\begin{equation}
    \mathrm{E}_\mathrm{rep}(\mathcal{G}_0, x) = \min_{z_0 \in \mathbb{R}^{k_0}} \| x - \mathcal{G}_0(z_0) \|.
\end{equation}
The following theorem from \cite{bora} gives an upper bound on overall recovery error:

\begin{theorem} \citep{bora}
Let $\mathcal{G}_0: \mathbb{R}^{k_0} \mapsto \mathbb{R}^n $ be an $L$-Lipschitz function (such as a neural network). Let $A \in \mathbb{R}^{m \times n}$ be a random Gaussian matrix for $m = O(k_0 \log \frac{Lr}{\delta})$, scaled so that $A_{ij} \sim \mathcal{N}(0, 1/m)$. For any $x^* \in \mathbb{R}^n$ and any observation $y = A \: x^* + \eta$, let $\hat{z}_0$ minimize $\| y - A \: \mathcal{G}_0(z_0)\|$ to within additive $\epsilon$ of the optimum over vectors with $\| z_0 \| \le r$. Then with probability $1 - e^{-\Omega(m)}$,
\begin{align}
    \|x^* - \mathcal{G}_0(\hat{z}_0) \| & \le 
        6 \min_{\|z_0\| \le r} \mathrm{E}_\mathrm{rep}(\mathcal{G}_0, x^*) \nonumber \\
        &\qquad\qquad 
        + 3 \| \eta \| + 2 \epsilon + 2 \delta.
\end{align}
\label{thm:bora-theorem1.2}
\vspace{-\baselineskip}
\end{theorem}
Thm.~\ref{thm:bora-theorem1.2} shows that, given sufficient measurements $m=O(k_0 \log {L})$, the error in the recovered image is bounded with high probability by the representation error of the model, the norm of the measurement noise $\eta$, optimization error $\epsilon$, and additive slack $\delta$ controlled by the number of measurements. 


We discuss the conditions required to apply Thm.~\ref{thm:bora-theorem1.2} and directly compare the recovery error of a cut and uncut generator on the same target image $x$. 
First, let $L = \max\{\|\mathcal{G}_0 \|_{lip}, \| \mathcal{G}_c \|_{lip}\}$. As $k_c > k_0$, setting $m=O \left(k_c \log \frac{Lr}{\delta} \right)$ is sufficient for Thm.~\ref{thm:bora-theorem1.2} to hold for both $\mathcal{G}_0$ and $\mathcal{G}_c$. As in \citet{bora}, we assume that both models can be optimized to the same level of error. 


Under these conditions, measurement error $\| \eta \|$, optimization error $\epsilon$, and the additive slack $\delta$ are the same for both $\mathcal{G}_0$ and $\mathcal{G}_c$. 
Then the difference in recovery error depends only their representation errors.
\begin{align}
\label{eqn:lower-repr-error}
    \mathrm{E}_{rep}(\mathcal{G}_c, x) & = \min_{z_c \in \mathbb{R}^{k_c}} \| x - \mathcal{G}_c(z_c) \| \nonumber \\
    & \le \min_{z_0 \in \mathbb{R}^{k_0}} \| x - \mathcal{G}_c(B_{c-1} \circ \ldots \circ B_0 (z_0)) \| \nonumber\\
    & = \min_{z_0 \in \mathbb{R}^{k_0}} \| x - \mathcal{G}_0(z_0) \| = \text{E}_{rep}(\mathcal{G}_0, x)
\end{align}
where in the inequality Eqn.~\eqref{eqn:lower-repr-error}, we have used the fact that the range of $B_{c-1} \circ \ldots \circ B_0(z_0)$ is a subset of $\mathbb{R}^{k_c}$. 



Intuitively, the range of $G_0$ is a $k_0$-dimensional manifold while the range of $G_c$ is a $k_c$-dimensional manifold. The representation error is the error induced by projecting the true image onto either of these manifolds. Our experiments show that representation error is the dominant factor of total recovery error and that it can be significantly reduced with access to the extra degrees of freedom provided by \ours{}. While the above analysis requires $m$ to be proportional to $k_c$, we find empirically that cutting \layers{} greatly improves performance, even for very few measurements.

\section{Related Work}
\label{sec:related-work}

There exists a substantial amount of prior work on the use of generators as signal priors.
\citet{creswell2016} use a fixed, trained generator, and optimize a logistic loss function in order to approximately recover a target image without compressed sensing.
\cite{bora} investigate the use of generative models as signal priors for compressed sensing and demonstrate that they can outperform sparsity priors at low undersampling ratios. They also provide theory describing the sources of error and show that representation error was a key limiting factor in recovery quality.

Methods for image recovery using generator signal priors may optimize latent code, model weights, or both.
\cite{mgan-prior} recover a target image by using multiple latent codes, passing each code through the first part of a network, taking a linear combination at an intermediate layer using tunable weights, and then passing that combined representation through the second part of the network.
This approach improves image quality, but requires substantial hyperparameter tuning and a complicated optimization procedure.
Unlearned methods such as Deep Decoder \citep{deep-decoder} and Deep Image Prior \citep{deep-image-prior} use no training data, and instead recover an image by optimizing the weights of a randomly-initialized network. 
These methods show that neural net architectures have structural bias towards natural images.
\citet{iagan} train a generator by standard methods, and then for a given target image, optimize both the latent code \textit{and} the generator's weights. 
This approach obtains some benefits from both learned and unlearned approaches and gives impressive results by reducing representation error. However, as there are many optimized parameters at test time, it results in a costly optimization procedure for each image.

Other methods have used specialized architectures and/or hybrid approaches to improve image recovery, but are computationally expensive or complicated.
\citet{inns} use invertible neural networks to obtain a high-dimensional latent representation with zero representation error by design, though these networks are extremely expensive both to train and during image recovery, taking 11 GPU minutes to recover a single $128 \times 128$ image.
In \cite{daniels2020reducing}, a linear combination of the output of a GAN with the output of a Deep Decoder model is used to recover an image.
This method is able to improve performance but requires an optimization that is more expensive than either a simple GAN-based recovery or a standard Deep Decoder recovery.
\cite{lcm} focus on increasing latent dimension to reduce representation error. They train a shared generator with high-dimensional input, as well as an a single latent convolutional net (similar to a Deep Image Prior) for every image, resulting in an expensive and complicated optimization.

\section{Experiments}
\label{sec:experiments}
We illustrate the effect of \ours{} on various architectures on compressed sensing as described in Sec.~\ref{sec:method}. We follow \citet{bora} and set the standard deviation of the noise $\eta$ such that $\sqrt{\mathbb{E} \left[\|\eta\|^2\right]} = 0.1$ for $64$px images, and scale the noise level for larger images such that $\mathbb{E} \left[\|\eta\|^2/ \|Ax\|^2\right]$ is kept constant. We compare against the standard sparsity-based Lasso-DCT to demonstrate utility. Lasso-DCT solves $\hat{z} = \min_{z} \| y - A \Phi z\|^2 + 0.01 \| z \|_1$, where $\Phi$ is the inverse discrete cosine transform matrix and the estimated image is $\hat{x} = \Phi \hat{z}$ \citep{lasso}. We find that DCT performs slightly better than a wavelet basis. 

We then compare \ours{} against several recent learned and unlearned methods - IAGAN, Deep Decoder (DD), and mGANprior - to assess its relative performance. See Sec. \ref{sec:related-work} for more details about these methods.
Note that our goal in these experiments is not to produce the highest PSNR possible (as GS does not optimize generator weights), but to show that this simple algorithm is competitive with recent methods.

\subsection{Model Architectures and Datasets Design}
We apply \ours{} (GS) to DCGAN \citep{dcgan}, BEGAN \citep{began}, and VAE \citep{vae} architectures with image sizes of 64px, 128px, and 128px, respectively. All models were trained on CelebA \citep{celeba}; see Appendix for generated samples. Note our method works for any pretrained generator. We measure recovery performance for compressed sensing (CS) using average peak signal-to-noise ratio (PSNR).
For each architecture, we choose the cut index $c$ that maximizes average PSNR over $100$ validation images on CelebA. 

We evaluate each architecture on test images from CelebA (``in-training-distribution'') and from COCO 2017 (``out-of-training-distribution'', \citet{coco}). For all experiments, we use $100$ images per dataset and $R=3$ restarts. Additional details are in the Appendix.

\begin{figure*}[t]
\begin{center}
\includegraphics[width=\textwidth]{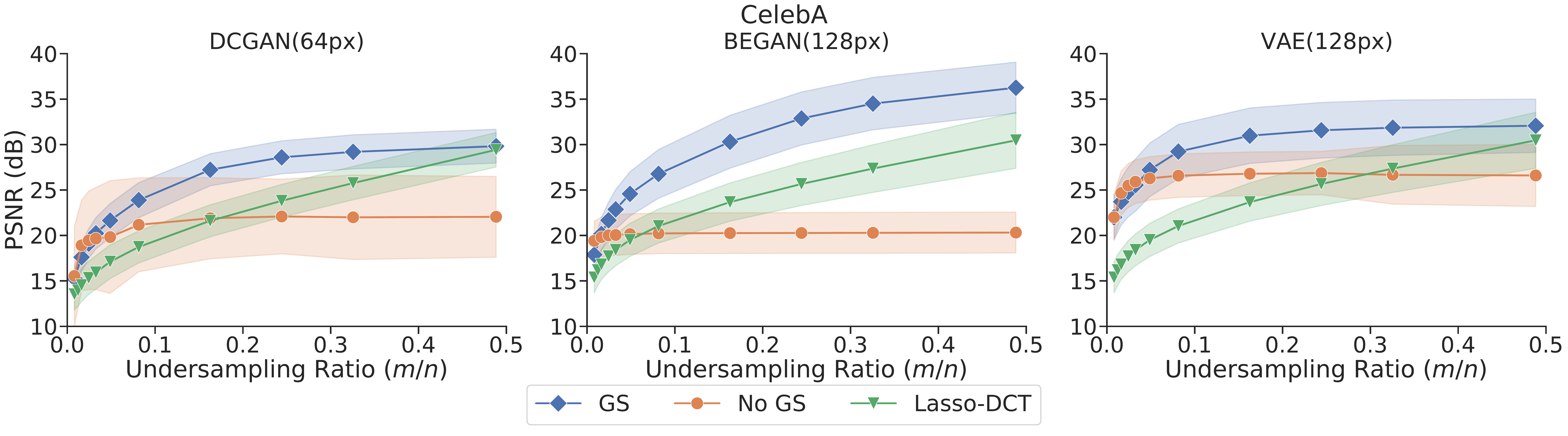}
\end{center}
\vspace*{-\baselineskip}
\caption{Effect of \ours{} for DCGAN (64px), BEGAN (128px), and VAE (128px) on CS of CelebA test-set images (all models were trained on CelebA). Shaded bands show $1$ standard deviation. GS greatly improves performance and outperforms Lasso-DCT.}
\label{fig:cs-celeba}
\end{figure*}

\begin{figure*}[!t]
\begin{center}
\includegraphics[width=\textwidth]{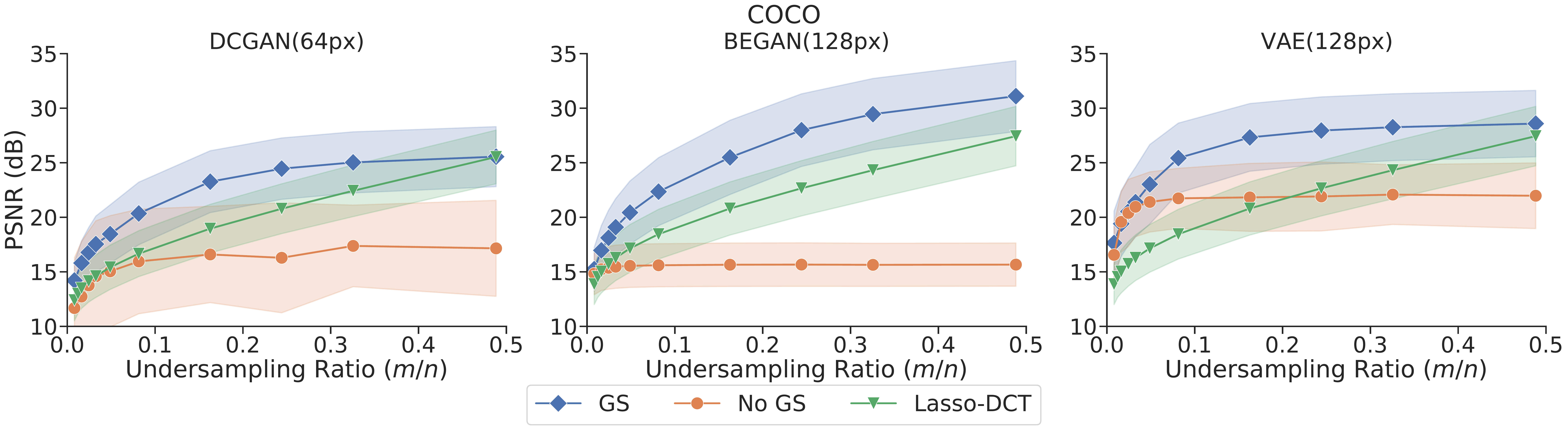}
\end{center}
\vspace*{-\baselineskip}
\caption{Effect of \ours{} for DCGAN (64px), BEGAN (128px), and VAE (128px) on CS of COCO images. Shaded bands show $1$ standard deviation. GS greatly improves performance even for ``out-of-training-distribution'' images.}
\label{fig:cs-ood}
\end{figure*}

\subsection{Effect of Generator Surgery}

\paragraph{In-training-distribution.} Fig.~\ref{fig:cs-celeba} shows the recovered PSNR averaged over test images from CelebA.
No GS replicates the experiments of \citet{bora}, and is able to outperform the Lasso-DCT baseline at low measurement regimes for all models. We see that GS improves recovery performance substantially for all architectures, and outperforms the Lasso-DCT baseline over a larger range of undersampling ratios. 

\paragraph{Out-of-training-distribution.}
In Fig.~\ref{fig:cs-ood}, we perform the same experiments using images from the COCO dataset. Even on these ``out-of-training-distribution'' images, cutting blocks gives a large jump in performance and our method outperforms Lasso-DCT almost everywhere, whereas No GS models (with the exception of VAE) do not. 
The performance increase over No GS is similar to CelebA images. 
This may indicate that much of the bias towards the training domain is contained in the early \layers{} of the generator. 

\paragraph{SOTA architecture.}
We apply \ours{} to a state-of-the-art class-conditional GAN (BigGAN \cite{biggan}) with an output image size of 512px. At this size, Gaussian compressed sensing exceeds reasonable memory constraints, so we consider other typical inverse problems (random inpainting and super-resolution). \ours{}~improves image recovery performance considerably ($\sim 10$dB, see Appendix for results).

\subsection{Comparison to Generator Priors}
\begin{figure}[t]
\begin{center}
\includegraphics[width=\columnwidth]{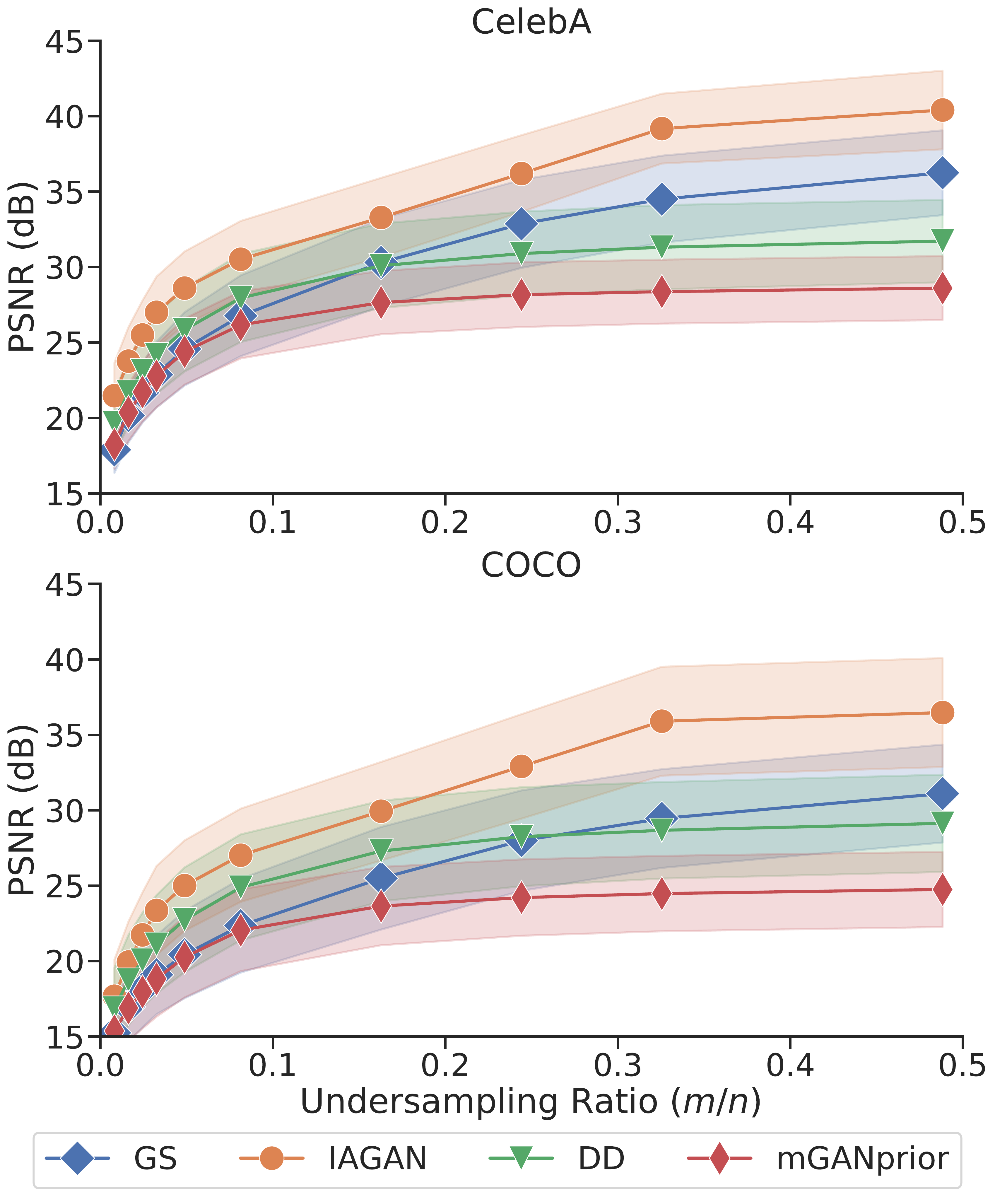}
\end{center}
\vspace*{-\baselineskip}
\caption{Baseline comparisons on CS of CelebA and COCO images for BEGAN(128px). All methods use 3 random restarts. Our method GS outperforms mGANprior and performs similarly to Deep Decoder (DD).}
\label{fig:cs-baselines}
\end{figure}

To compare \ours{} against generator prior baselines, we use the same generator architecture for both \ours{} and for IAGAN and mGANprior. Deep Decoder (DD) uses its own architecture and we use the optimized implementation from \cite{inns} at both 64px and 128px. We change the number of filters to match the number of tunable parameters as closely as possible to our method. We generally follow the hyperparameters given in the literature, except for mGANprior, which had to be individually tuned as the original settings were not applicable. All baselines were also run with 3 random restarts for a fair comparison.

\begin{table}[b]
\centering
\begin{threeparttable}
\caption{Overparameterization ratio}
\begin{tabular}{lrrr}
\hline
             & DCGAN & BEGAN & VAE   \\ \hline \hline
mGANprior    & 0.123 & 0.195 & 0.250 \\
\textbf{GS}  & 0.667 & 0.833 & 0.667 \\
DD\tnote{*}  & 0.684 & 0.838 & 0.838 \\
IAGAN        & 291   & 43.7  & 855   \\ \hline
\end{tabular}
\begin{tablenotes}\scriptsize
\item[*] DD uses fixed architectures for 64px and 128px
\end{tablenotes}
\label{tab:overparam-ratio}
\end{threeparttable}
\end{table}

\begin{figure}[t]
\begin{center}
\includegraphics[width=\columnwidth]{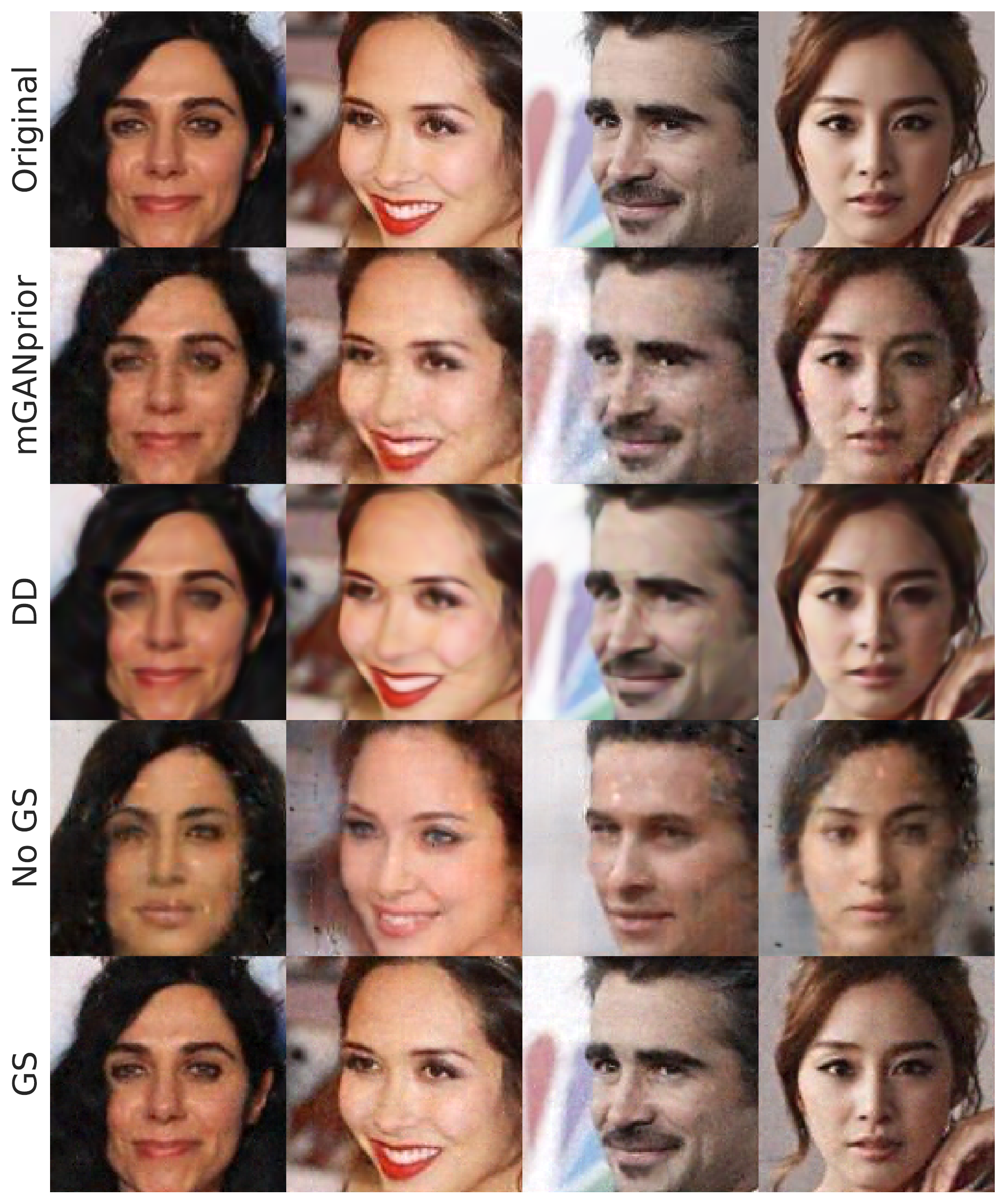}
\end{center}
\caption{Compressed sensing of CelebA images for BEGAN(128px) at $m/n\approx 0.16$. No GS produces clearly different faces, exhibiting high representation error. DD and GS perform similarly, but DD smooths over small details (hair/beard strands, individual teeth).} 
\label{fig:cs-images-celeba}
\vspace*{-\baselineskip}
\end{figure}

\begin{figure}[t]
\begin{center}
\includegraphics[width=\columnwidth]{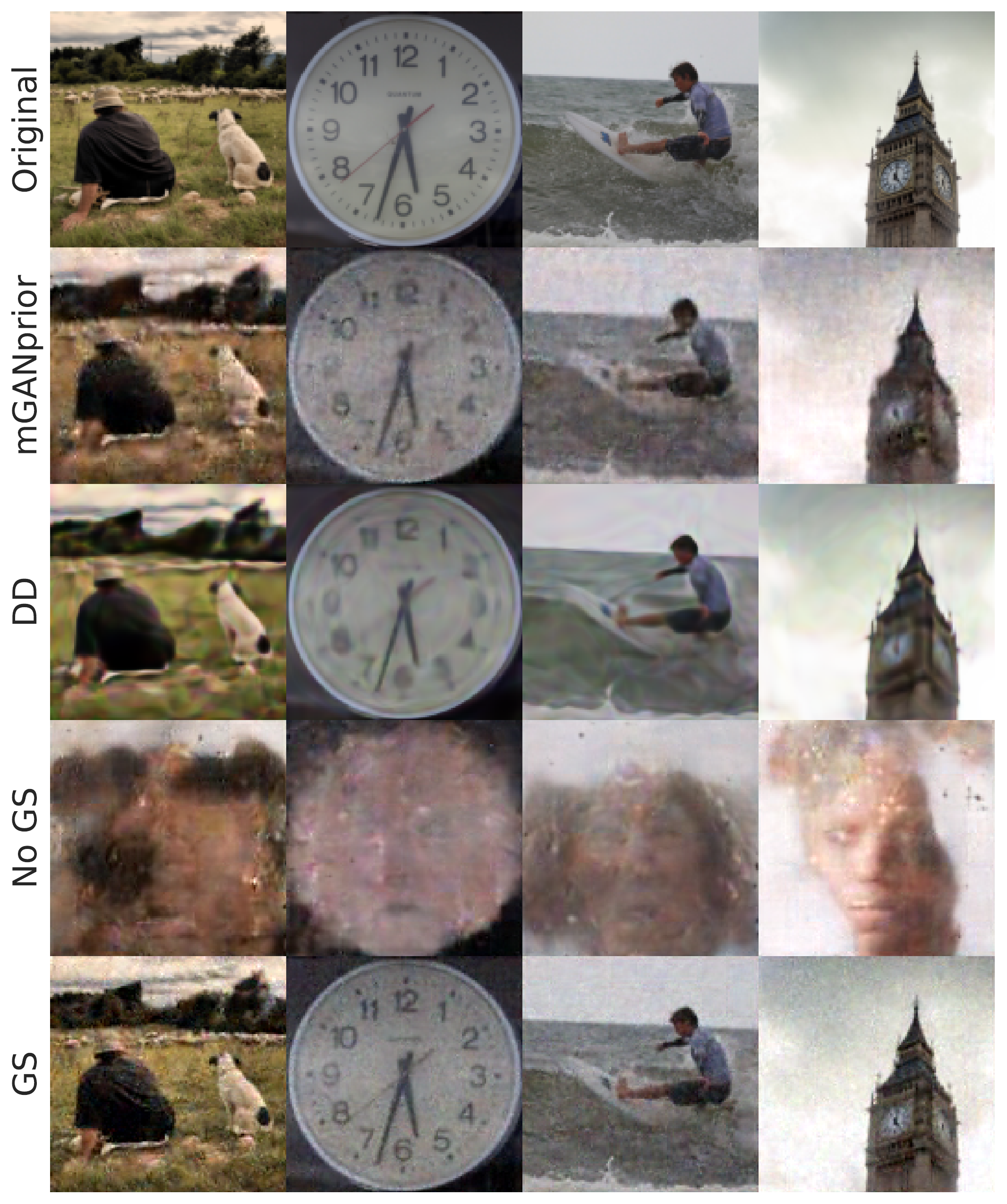}
\end{center}
\caption{Compressed sensing of COCO images for BEGAN(128px) at $m/n\approx 0.16$. No GS creates non-existent face-like artifacts as it was trained on celebrity faces. DD and GS perform similarly, but DD smooths over small details (clock numerals, water texture).}
\label{fig:cs-images-coco}
\vspace*{-\baselineskip}
\end{figure}

Fig.~\ref{fig:cs-baselines} compares \ours{} with baselines for BEGAN (128px) and Table ~\ref{tab:overparam-ratio} lists the overparameterization ratio of each method, i.e. the ratio of the number of tunable parameters to the number of pixels in a target image. The comparisons for other architectures and other details are provided in the Appendix.
We emphasize that our focus is to provide a conceptually method that is competitive with recent baselines. GS generally beats mGANprior and performs similarly to DD: it is slightly worse at lower measurements and slightly better at higher measurements. 
IAGAN outperforms all methods by a fair margin. This is expected because IAGAN is significantly overparameterized; as a general trend, performance increases with the number of optimizable parameters.

Figs. ~\ref{fig:cs-images-celeba} and ~\ref{fig:cs-images-coco} show recovered samples from CelebA and COCO for compressed sensing at $m/n \approx 0.16$ for BEGAN and baselines. 
First, we can clearly see that No GS recovers a different face than the original for CelebA and hallucinates a non-existent face for COCO. This is expected as the models are trained on CelebA, and illustrates the large representation error in the absence of surgery. 
mGANprior produces slightly lower quality results than GS; it is grainier and sometimes creates different colors than the original. 
DD and GS show subtle qualitative differences. 
DD often produces smoother images, sometimes losing fine textural details such as eyes, hair strands, and clock numerals;
GS produces grainier images but recovers such fine details.
\section{Understanding the Effect of Surgery}
\label{sec:empirical-studies}

In this section, we perform experiments to address several key questions about \ours{}. We confirm that representation error is reduced after surgery and that more computation for the uncut generator does not help. We also demonstrate that using pre-trained weights is required for \ours{} and that surgery can only be applied at inversion time.

\subsection{Is Representation Error Reduced?}

\paragraph{Optimization error.}
We showed in Sec.~\ref{sec:method} that cutting \layers{} can reduce representation error. 
To support this experimentally, we first eliminate the error $\delta$ by setting $A=I$ and eliminate measurement error by setting $\eta=0$. 
In this case, the only remaining sources of error are representation error and optimization error. 

\begin{table}[b]
\caption{PSNR on train and generated images}
\label{tab:repr-error}
\begin{center}
\begin{tabular}{lrr}
\hline
          & \multicolumn{1}{c}{No GS ($\mathcal{G}_0$)} & \multicolumn{1}{c}{\textbf{GS} ($\mathcal{G}_c$)}       \\ \hline
Train     & $20.51\text{dB} (\pm 2.12)$ & $42.25\text{dB} (\pm 2.65)$ \\
Generated & $55.65\text{dB} (\pm 2.68)$ & $45.14\text{dB} (\pm 0.59)$ \\ \hline
\end{tabular}
\end{center}
\vspace*{-\baselineskip}
\end{table}

We compare the performance of GS and No GS in this setting for the BEGAN architecture on both train and generated images (Table.~\ref{tab:repr-error}).
Generated images are randomly sampled from the uncut generative model by taking $x^* = \mathcal{G}_0(z_0^*), z_0^* \sim \mathcal{N}(0,1)$.
In this case, the representation error for $\mathcal{G}_0$ is: $\text{E}_{rep}(\mathcal{G}_0, x^*) = \min_z \| x^* - \mathcal{G}_0 (z)\| = \|x^* - \mathcal{G}_0(z_0^*)\| = 0$.
Note that the range of $\mathcal{G}_c$ is a superset of the range of $\mathcal{G}_0$, so that its representation error is also zero.
For these \mbox{``Generated''} images, we see that the uncut generator has a very high PSNR and thus very small optimization error.
However, the PSNR is quite low for training images ($\approx 20$dB), confirming that $G_0$ suffers from representation error.
GS performs slightly worse on the same ``Generated'' images, indicating it has a somewhat higher optimization error.
Since GS achieves nearly the same quality on training and generated images, we deduce that it has greatly reduced representation error.

\begin{figure}[t]
\begin{center}
\includegraphics[width=\columnwidth]{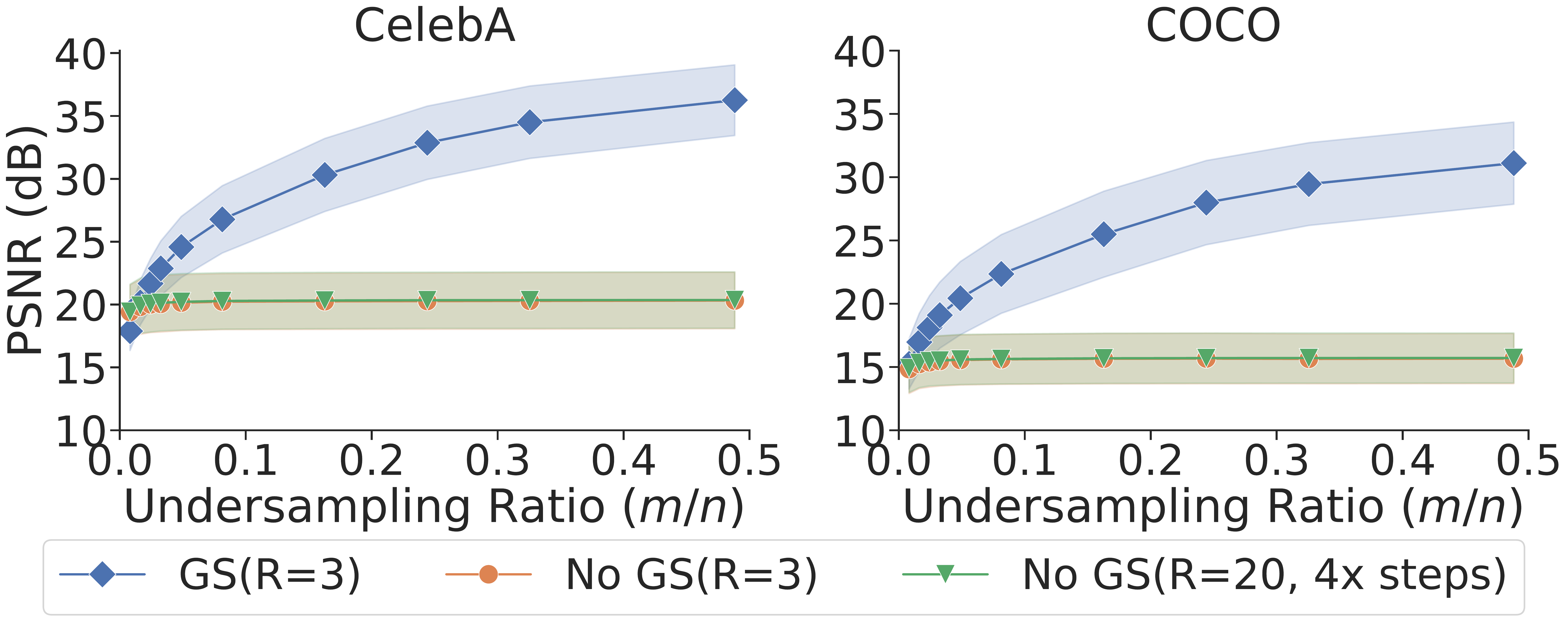}
\end{center}
\vspace*{-\baselineskip}
\caption{GS vs. No GS for BEGAN(128px) on CS of images from CelebA and COCO. The No GS lines appear overlapped. The performance of No GS does not improve from using $6$x more restarts and $4$x more optimization steps.}
\label{fig:restarts}
\end{figure}

\paragraph{More computation.}
To further demonstrate that optimization error is not a factor, we increase the computational budget for the No GS model by using $6$x more restarts and $4$x more gradient descent steps. We compare this against our method on compressed sensing, with results shown in Fig.~\ref{fig:restarts}. We observe that more computation does not improve the quality of the uncut generator, providing further evidence that our image recovery algorithm converges to a global optimum.

\paragraph{Other architectures.}
\begin{figure}[t]
\begin{center}
\includegraphics[width=\columnwidth]{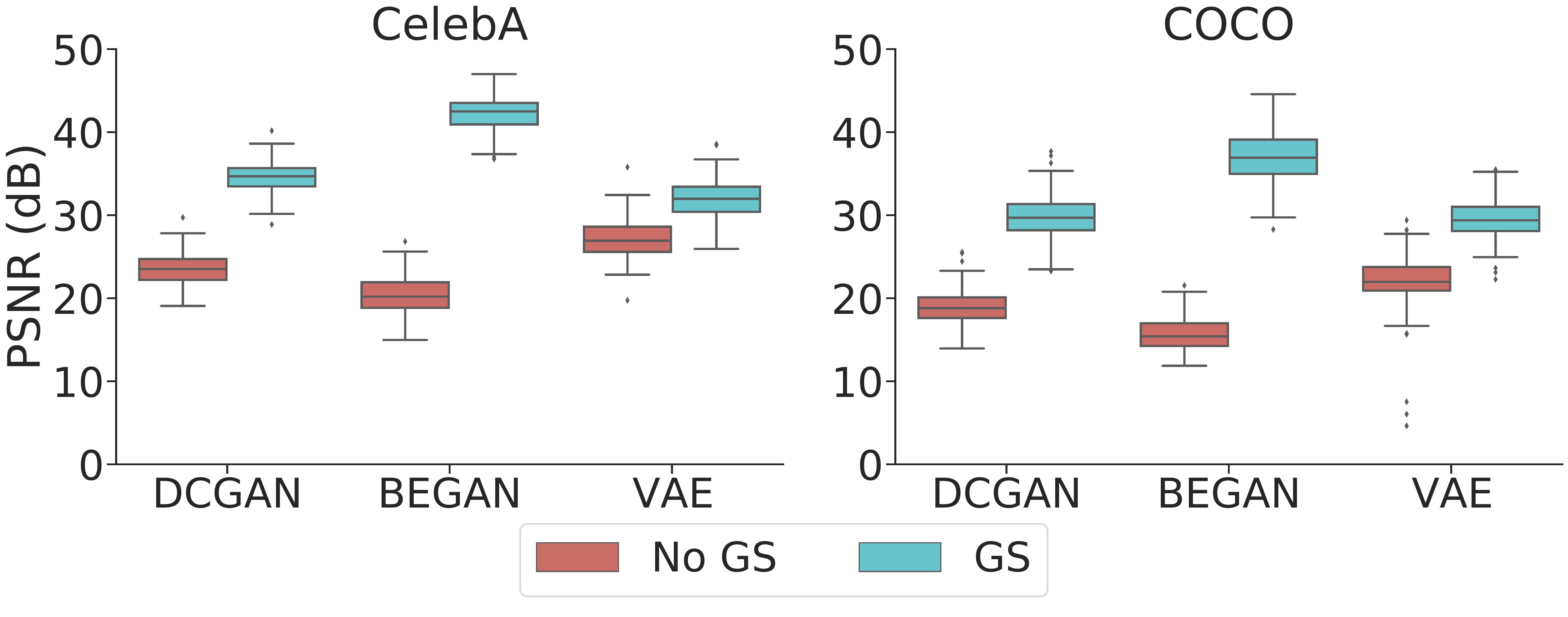}
\end{center}
\vspace*{-\baselineskip}
\caption{Effect of GS on image reconstruction of images from CelebA and COCO. The difference in PSNRs represent approximately how much representation error has decreased with surgery.}
\label{fig:noop}
\end{figure}

We examine the change in representation error for GS on different architectures. We again study the case where $A=I$ and $\eta = 0$. Fig.~\ref{fig:noop} compares \ours{} with No GS for DCGAN, BEGAN, and VAE on test images from CelebA and COCO. Our method benefits all architectures, reducing representation error and producing higher quality reconstructions ($\approx 10$dB higher).

\subsection{Do Learned Weights Matter?}
\begin{figure}[t]
\begin{center}
\includegraphics[width=\columnwidth]{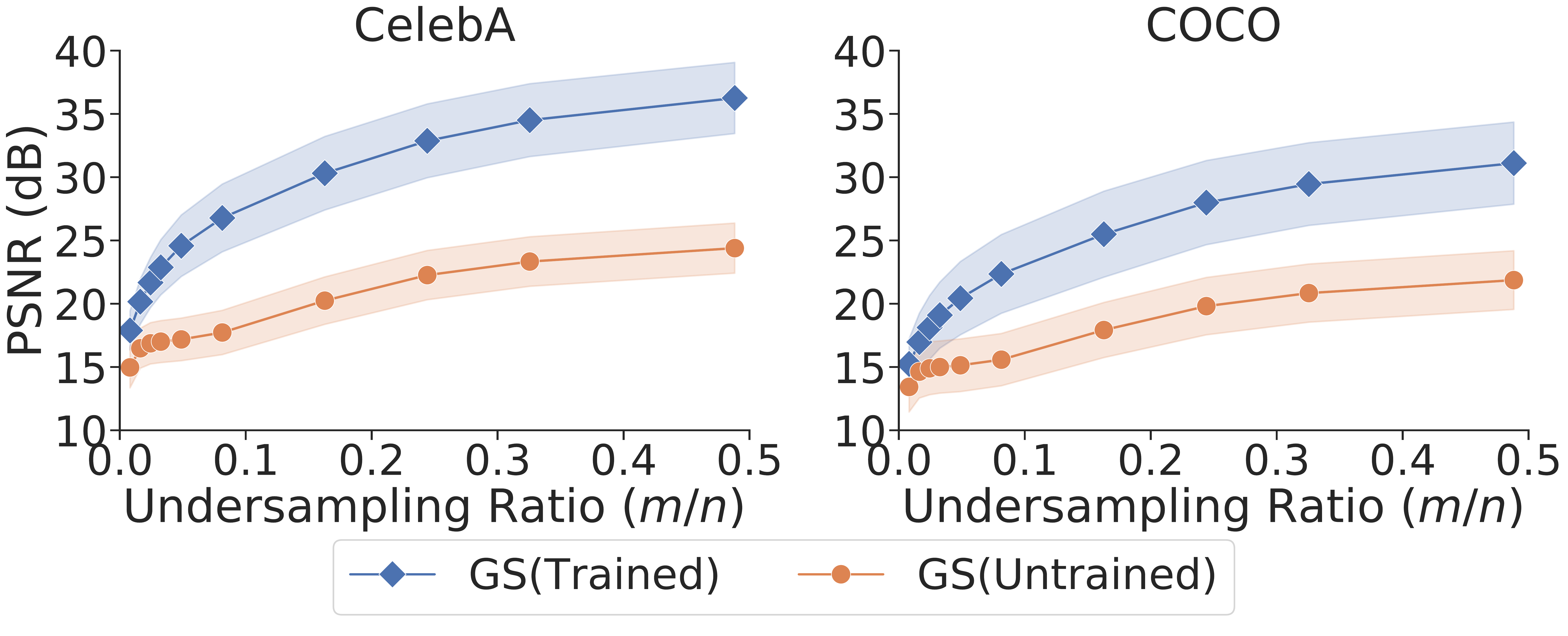}
\end{center}
\caption{Comparison of trained vs. untrained GS for BEGAN on CS of CelebA and COCO images. Using pre-trained weights increases performance considerably.}
\label{fig:untrained}
\end{figure}
As discussed in Sec. ~\ref{sec:method}, \ours{} effectively trades the generative capability for a higher dimensional latent representation, which leads to increased performance. A natural question to ask is whether the performance increase is solely due to the increased degrees of freedom and whether the learned weights are required at all.

Fig. ~\ref{fig:untrained} compares recovery performance using a trained BEGAN generator or using a randomly initialized generator in Alg.\ref{alg:image-recovery}. We can see that the trained model weights produce significantly higher recovery quality. 

\subsection{Can GS be Used Before Training?}
We have shown that early \layers{} of several architectures can be cut to improve image recovery, which leads to the question of whether these \layers{} are necessary at all.
In other words, we might try to simply train a model with a high-dimensional latent code, and directly use this model for inversion. 
We attempt to train DCGAN and BEGAN with $c=1$ cuts for a variety of hyperparameter settings and train for at least $25$ epochs in each case. 
All training scenarios fail within the first epoch and suffer mode collapse (see Appendix). 
This observation is consistent with our expectations, and supports the notion that test-time modifications are required for \ours{}.

\subsection{Other considerations}
\paragraph{How to choose cut index.} For each model architecture, we choose the cut index $c$ that maximizes mean PSNR on a set of validation images from CelebA (see Appendix). 
There is a tradeoff in choosing how many \layers{} to cut. 
Cutting the initial \layers{} increases the dimension of the intermediate representation and therefore increases expressivity of the generator. 
It also relaxes the inductive bias of the model towards the training domain. 
However, as the cut index $c$ approaches $d$, the cut generator loses too much of its learned weights and cannot generalize to natural images from measurements.

\paragraph{Initialization strategies.}
As \ours{} greatly increases the latent dimension, regularization is required for stable optimization. 
We claim that the initialization of the latent vector in Alg.~\ref{alg:image-recovery} provides some of this regularization, and we compare the performance of different initialization strategies on BEGAN (see Appendix). 
We find that performance varies widely across different initialization strategies, and we select a censored normal distribution for all experiments.
\section{Discussion}
\label{sec:discussion}






We have shown that \ours{} provides a simple and efficient mechanism to greatly improve reconstruction quality in compressed sensing for a wide variety of model architectures. While uncut generators underperform relative to sparsity baselines, cutting early \layers{} results in signal priors that outperform the sparsity baselines over a broad range of undersampling ratios. The fact that \ours{} can recover better solutions than the uncut generator even for low measurements may show that our optimization procedure benefits from implicit regularization. Modeling this implicit regularization could strengthen the theoretical analysis of our method for future work.
Additionally, we observe high-quality reconstructions on images that are out-of-distribution relative to the dataset used to train the original generator. 
We observe that surgery cannot be performed before training to reduce representation error. We find simple initialization strategies that can successfully regularize our high-dimensional optimization.

Our results indicate that test-time architectural modifications can provide a substantial benefit for image recovery tasks. We note that \ours{} takes a generative model and outputs a signal prior that is no longer itself a generative model. That is, one cannot directly sample from the density of the training data using a cut generator after surgery. 
Conversely, the modified signal priors can represent more images than their underlying generative model. 
This increase in the range of the partial generator also implies that some images in the range will not be natural images.
In the context of solving inverse problems, the presence of unnatural images is acceptable because a sufficient number of measurements can exclude them being returned in a specific problem. 

We do not recommend \ours{} directly for other applications of generative modeling, such as data generation and data augmentation for training neural networks. That said, this work inspires future directions for generative modeling. For example, mixed optimizations such as IAGAN might succeed with drastically fewer optimizable parameters.  
Additionally, it may be possible to leverage our results for improved transfer learning of generative models.

\bibliography{references}

\appendix

\onecolumn
\aistatstitle{Generator Surgery: \\
Supplementary Materials}

\section{Experimental Setup}

\subsection{Model Architectures and Training Details}
We use an official PyTorch implementation of DCGAN\footnote{\url{https://github.com/pytorch/examples/tree/master/dcgan}} and implement BEGAN with skip connections based on the original paper. We use a standard convolutional VAE architecture\footnote{\url{ https://github.com/AntixK/PyTorch-VAE/blob/master/models/vanilla_vae.py}}. The cut index $c$ is chosen using a validation set (see Fig.~\ref{fig:best-cuts}).

For a state-of-the-art architecture, we use a pretrained BigGAN\footnote{\url{https://github.com/huggingface/pytorch-pretrained-BigGAN}} (which was trained on ImageNet). We evaluate its performance on $20$ test images from CelebA-HQ and COCO 2017. Note that both datasets are out-of-training distribution for this model. As BigGAN produces image sizes of 512px, compressed sensing experiments did not fit in GPU memory and we instead apply \ours{} to random inpainting (where the sensing matrix $A=I$ with a random subset of rows set to zero) and super-resolution ($A$ is the downsampling operator). 

All image inversions are performed on a single NVIDIA TITAN RTX.
We use L-BFGS as the optimizer for \ours{} as it performs slightly better than ADAM. Other hyperparameters are given in Table.~\ref{tab:gs-hyperparams}.

\begin{table}[ht]
\centering
\small
\begin{threeparttable}
\caption{Hyperparameters for \ours{}.}
\begin{tabular}{cll}
\hline
Model                          & Task                & Hyperparameters            \\ \hline
\multirow{2}{*}{DCGAN(64px)}   & CS                  & LR($z$) = $0.1, 25$ steps, $c=1$ \\
                               & Raw Reconstruction* & LR($z$) = $1, 100$ steps, $c=1$  \\ \hline
\multirow{2}{*}{BEGAN(128px)}  & CS                  & LR($z$) = $1, 25$ steps, $c=2$   \\
                               & Raw Reconstruction  & LR($z$) = $1, 100$ steps, $c=2$  \\ \hline
\multirow{2}{*}{VAE(128px)}    & CS                  & LR($z$) = $1, 25$ steps, $c=1$   \\
                               & Raw Reconstruction  & LR($z$) = $1, 40$ steps, $c=1$   \\ \hline
\multirow{2}{*}{BigGAN(512px)} & Inpainting          & LR($z$) = $1.5, 25$ steps, $c=7$ \\
                               & Super-resolution    & LR($z$) = $1.5, 25$ steps, $c=7$ \\ \hline
\end{tabular}
\begin{tablenotes}\footnotesize
\item[*] ``Raw reconstruction'' refers to reconstructing the image with no degradation ($A=I, \eta=0$).
\end{tablenotes}
\label{tab:gs-hyperparams}
\end{threeparttable}
\end{table}

\subsection{Generator Samples after Training}

\begin{figure}[t]
\centering
\begin{subfigure}{0.3\linewidth}
\includegraphics[width=\linewidth]{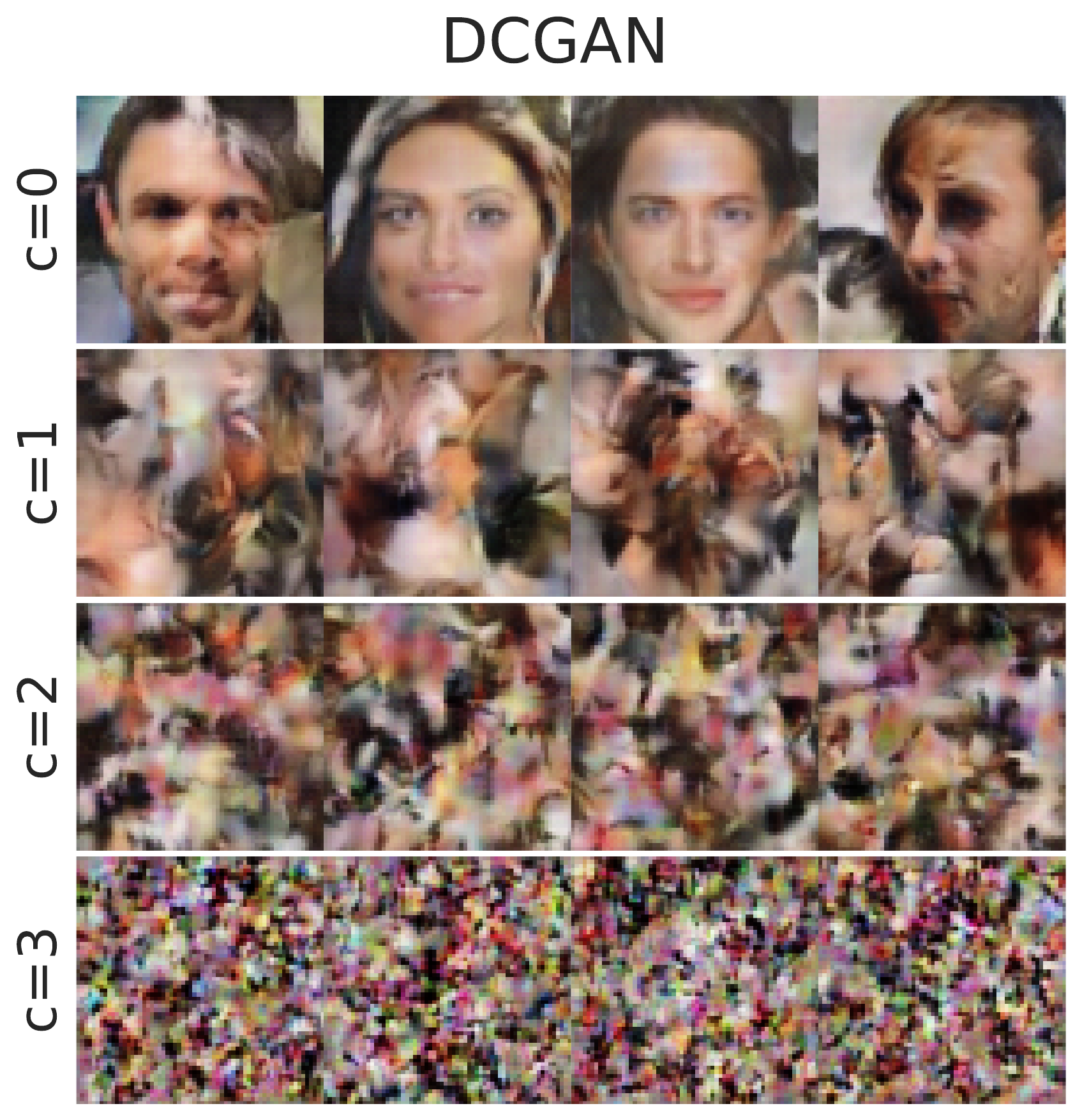}
\end{subfigure}
\begin{subfigure}{0.3\linewidth}
\includegraphics[width=\linewidth]{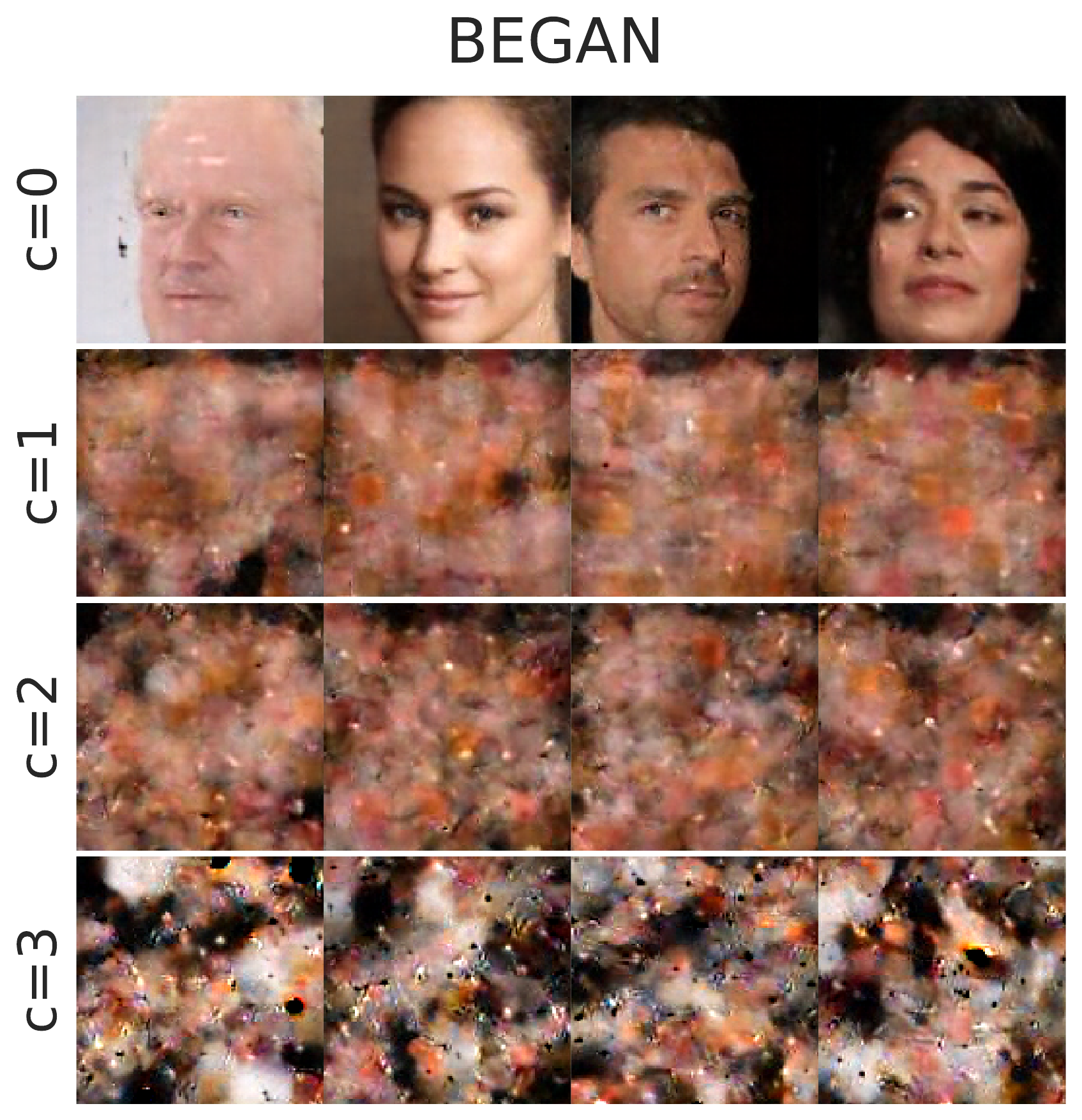}
\end{subfigure}
\begin{subfigure}{0.3\linewidth}
\includegraphics[width=\linewidth]{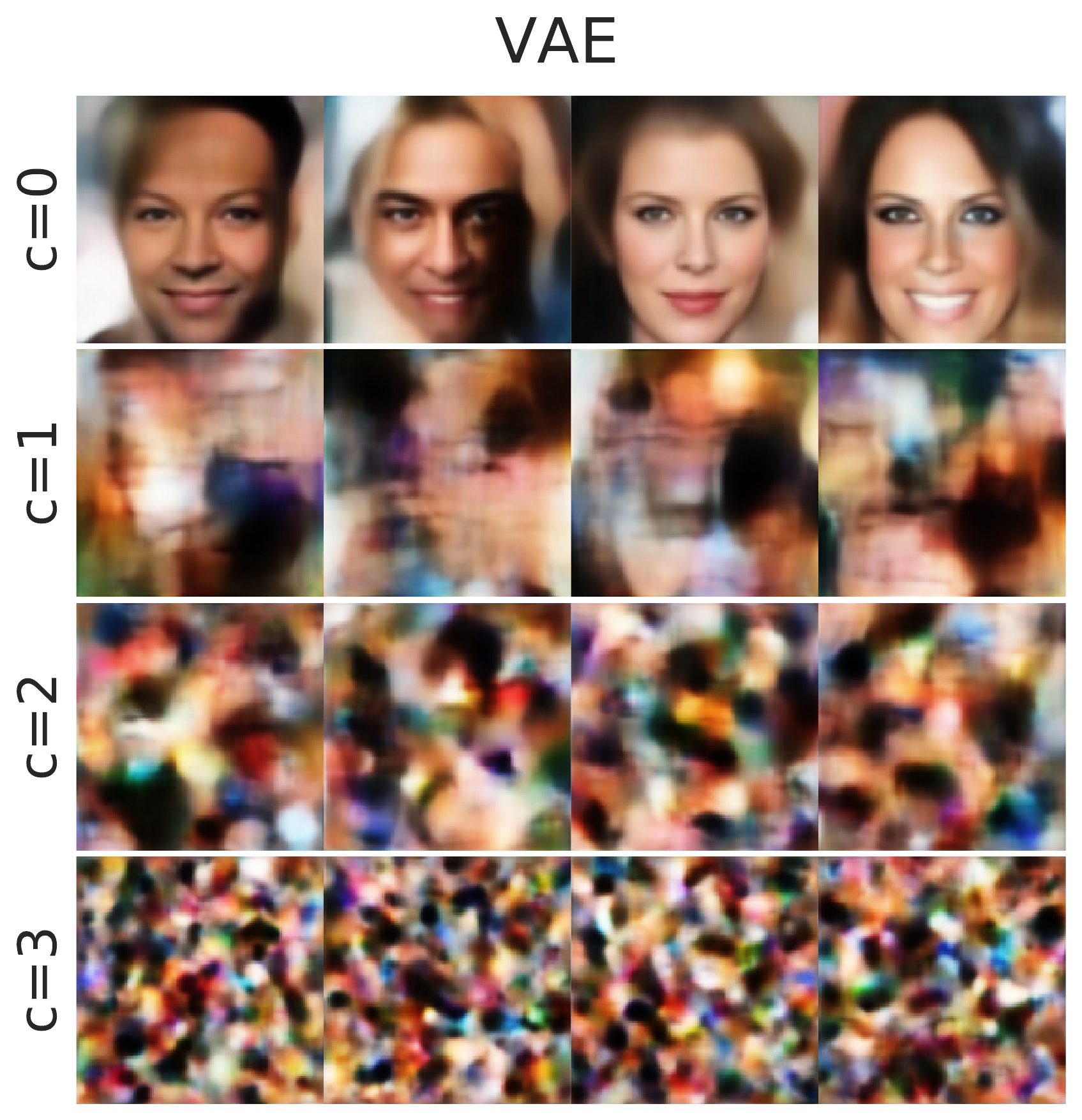}
\end{subfigure}
\caption{Generated samples from pre-trained DCGAN, BEGAN, and VAE with varying cut index.}
\label{fig:generator-samples}
\end{figure}

We give samples from our pre-trained generators, before \ours{} (``cuts$=0$"), and after cutting $1$, $2$, or $3$ \layers{} from each model. 
For all $c$, images are generated from latent codes sampled the same distribution. The high quality samples in the first rows show that our original generators have been successfully trained, while the low quality samples from the cut generators show that these no longer behave as a generative model after \ours{}.


\section{Experiments}

\subsection{SOTA Architecture}
Table ~\ref{tab:biggan} shows inpainting and super-resolution results on CelebA-HQ and COCO for both ``No GS'' and \ours{}. We see that cutting significantly improves performance even for a large state-of-the-art GAN, with almost a $10$dB increase in some cases. This shows that our method can be applied to a wide variety of pre-trained architectures and improves recovery performance not just for compressed sensing, but for other inversion problems as well.

\begin{table}[h]
\begin{center}
\caption{\ours{} on BigGAN(512px)}
\label{tab:biggan}
\begin{tabular}{llrr}
\hline
          &           & \multicolumn{1}{c}{No GS ($\mathcal{G}_0$)} & \multicolumn{1}{c}{\textbf{GS} ($\mathcal{G}_c$)}      \\ \hline \hline
CelebA-HQ & Inp (5\%)  & $16.08 (\pm 7.80)$ & $24.75 (\pm 7.24)$ \\
          & Inp (10\%) & $15.82 (\pm 7.87)$ & $24.66 (\pm 9.20)$ \\
          & SR (4x)   & $20.05 (\pm 3.58)$ & $23.16 (\pm 9.11)$ \\
          & SR (8x)   & $14.29 (\pm 7.47)$ & $21.65 (\pm 6.43)$ \\ \hline
COCO      & Inp (5\%)  & $13.08 (\pm 6.62)$ & $22.95 (\pm 3.38)$ \\
          & Inp (10\%) & $14.96 (\pm 5.80)$ & $24.21 (\pm 5.72)$ \\
          & SR (4x)   & $13.96 (\pm 5.68)$ & $20.88 (\pm 6.33)$ \\
          & SR (8x)   & $13.31 (\pm 5.51)$ & $16.18 (\pm 6.69)$ \\ \hline
\end{tabular}
\end{center}
\end{table}

\subsection{Comparison to Other Generator Priors}

\begin{table}[H]
\centering
\caption{Hyperparameters for Baseline Models for CS.}
\small
\begin{tabular}{cll}
\hline
Baseline                   & Architecture & Hyperparameters                                                                                                                  \\ \hline \hline
\multirow{3}{*}{mGANprior} & DCGAN(64px)  & LR($z$) = $3\mathrm{e}{-2}, 5000$ steps, latent codes $N = 10$, composing layer $l = 1$                                                             \\
                           & BEGAN(128px) & LR($z$) = $1\mathrm{e}{-3}, 3000$ steps, latent codes $N = 20$, composing layer $l = 2$                                                             \\
                           & VAE(128px)   & LR($z$) = $5\mathrm{e}{-3}, 3000$ steps, latent codes $N = 20$, composing layer $l = 1$                                                             \\ \hline
\multirow{5}{*}{IAGAN}     & DCGAN(64px)  & \begin{tabular}[c]{@{}l@{}}Stage 1: LR($z$) = $0.1, 1600$ steps\\ Stage 2: LR($z$) = $1\mathrm{e}{-4}$, LR($\theta$) = $1\mathrm{e}{-4}, 600$ steps\end{tabular} \\
                           & BEGAN(128px) & \begin{tabular}[c]{@{}l@{}}Stage 1: LR($z$) = $0.1, 1600$ steps\\ Stage 2: LR($z$) = $1\mathrm{e}{-4}$, LR($\theta$) = $1\mathrm{e}{-4}, 600$ steps\end{tabular} \\
                           & VAE(128px)   & \begin{tabular}[c]{@{}l@{}}Stage 1: LR($z$) = $0.1, 1000$ steps\\ Stage 2: LR($z$) = $1\mathrm{e}{-4}$, LR($\theta$) = $1\mathrm{e}{-4}, 600$ steps\end{tabular} \\ \hline
\multirow{2}{*}{DD}        & 64px         & LR($\theta$) = $1\mathrm{e}{-2}, 5000$ steps, depth $d = 5$, channels $k = 44$                                                                    \\
                           & 128px        & LR($\theta$) = $1\mathrm{e}{-2}, 5000$ steps, depth $d = 6$, channels $k = 89$                                                                    \\ \hline
\end{tabular}
\label{tab:baseline-hyperparams}
\end{table}

\begin{figure}[H]
\centering
\begin{subfigure}{0.8\linewidth}
\includegraphics[width=\linewidth]{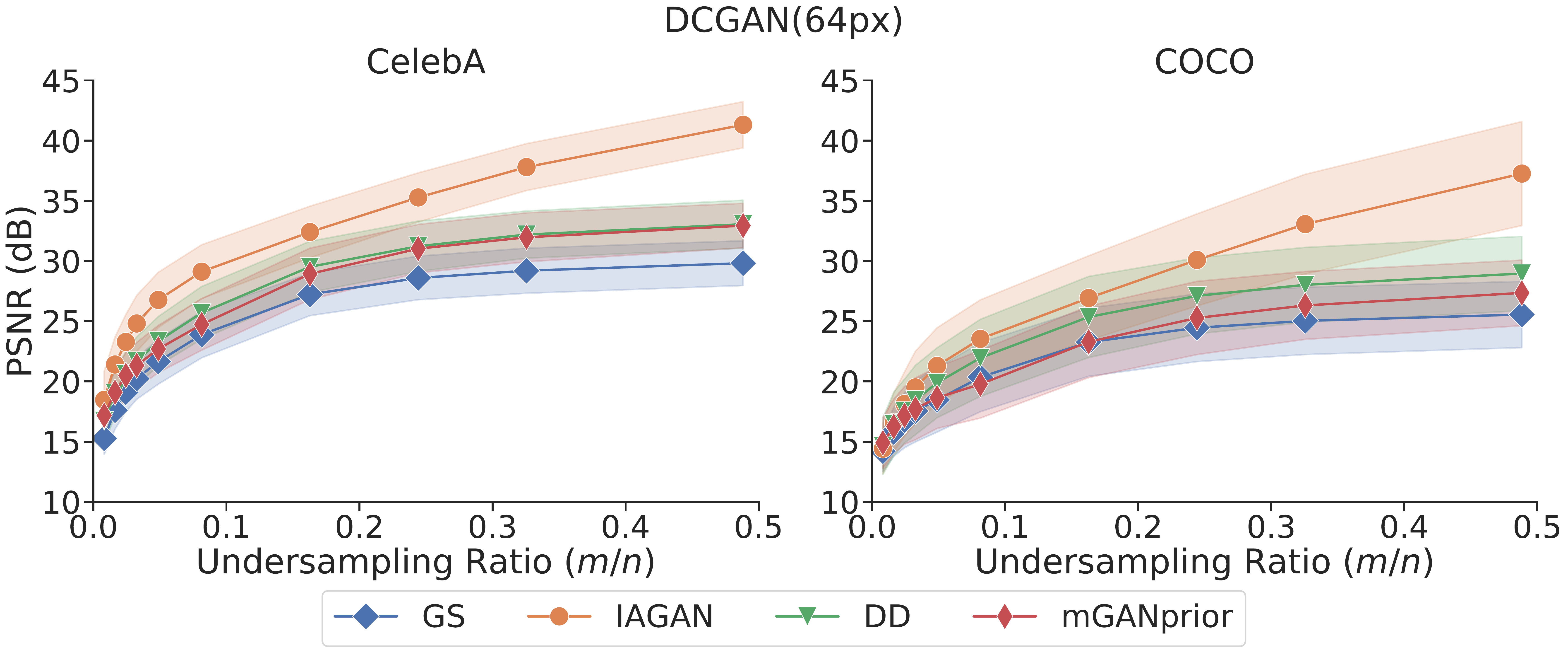}
\end{subfigure}
\begin{subfigure}{0.8\linewidth}
\includegraphics[width=\linewidth]{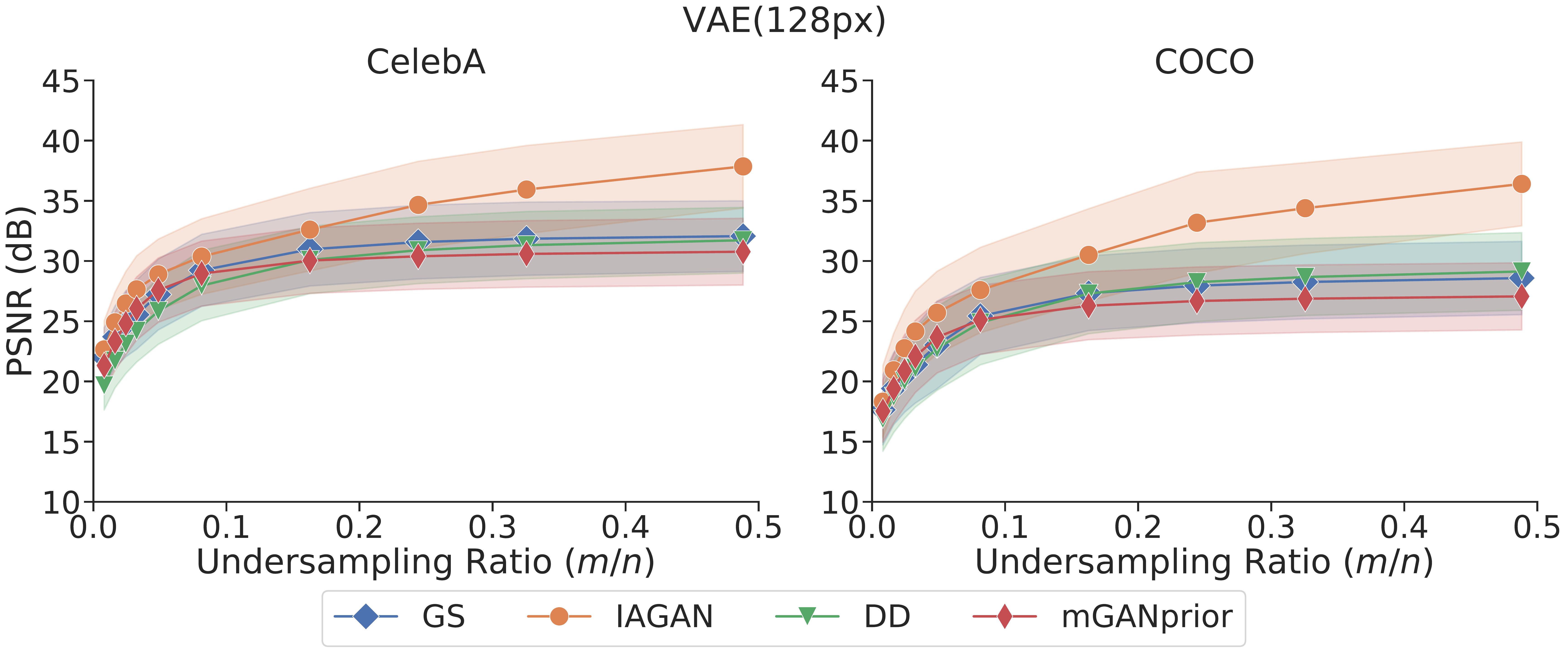}
\end{subfigure}
\caption{Baseline comparisons on CS of CelebA and COCO images for DCGAN(64px) and VAE(128px). All methods use 3 random restarts. Our method performs poorly for DCGAN but performs similarly to Deep Decoder (DD) for the VAE architecture.}
\label{fig:cs-baseline-supp}
\end{figure}

All generator prior baselines use the ADAM optimizer as in the original papers. Other hyperparameters are given in Table.~\ref{tab:baseline-hyperparams}.

We replicate the baseline comparisons performed in the main paper, using VAE and DCGAN generators in Fig.~\ref{fig:cs-baseline-supp}.
Here, we see that \ours{} performs worse for DCGAN, while VAE performs comparably to Deep Decoder.
We note that the performance of our method relies on the quality of the original generative model.
We hypothesize that our results can be improved even further by utilizing better generator architectures.

\newpage
\section{Effect of Surgery Results}

\subsection{Surgery before Training}

\begin{figure}[h]
\centering
\begin{subfigure}{0.35\linewidth}
\includegraphics[width=\linewidth]{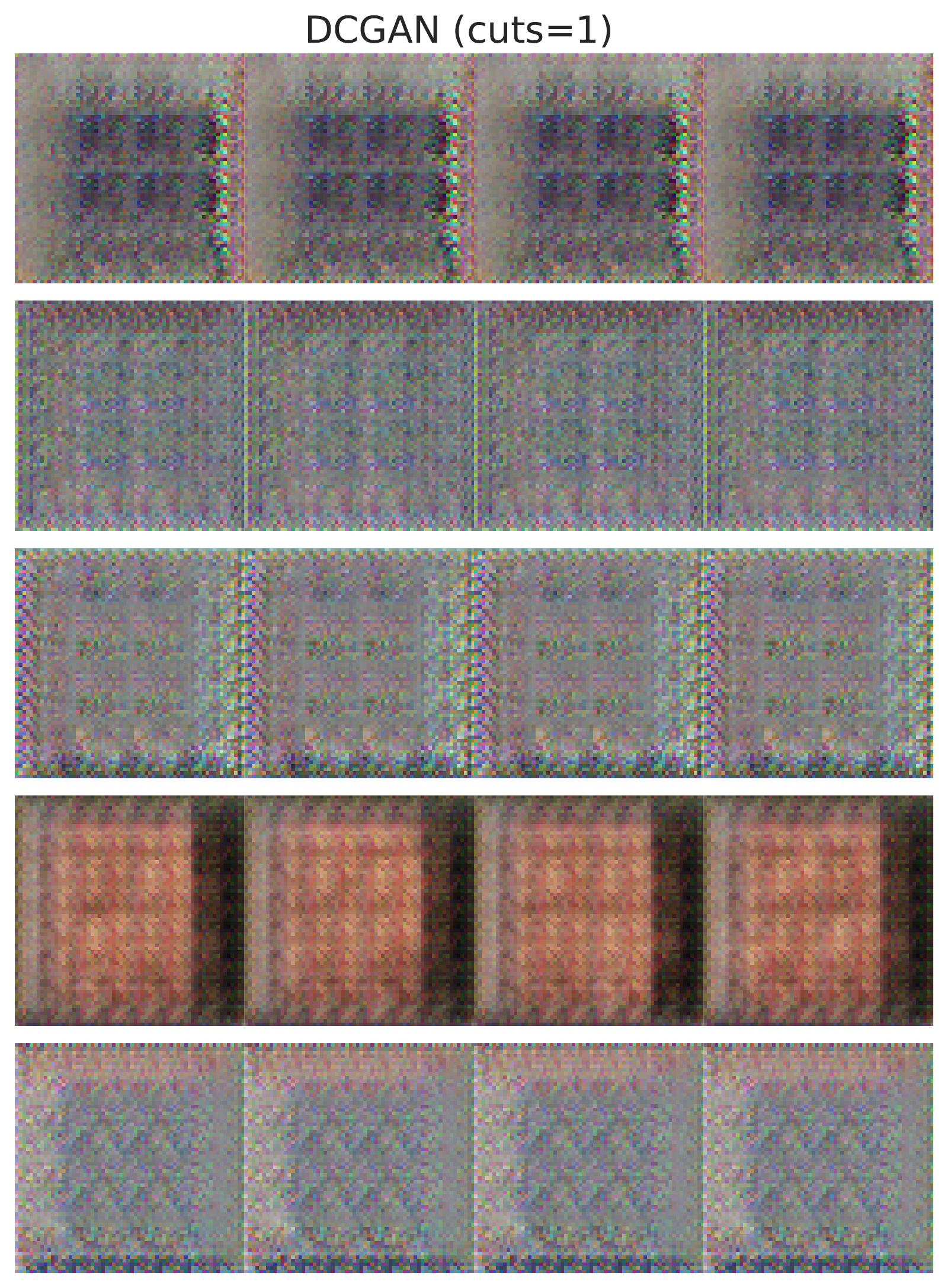}
\end{subfigure}
\begin{subfigure}{0.35\linewidth}
\includegraphics[width=\linewidth]{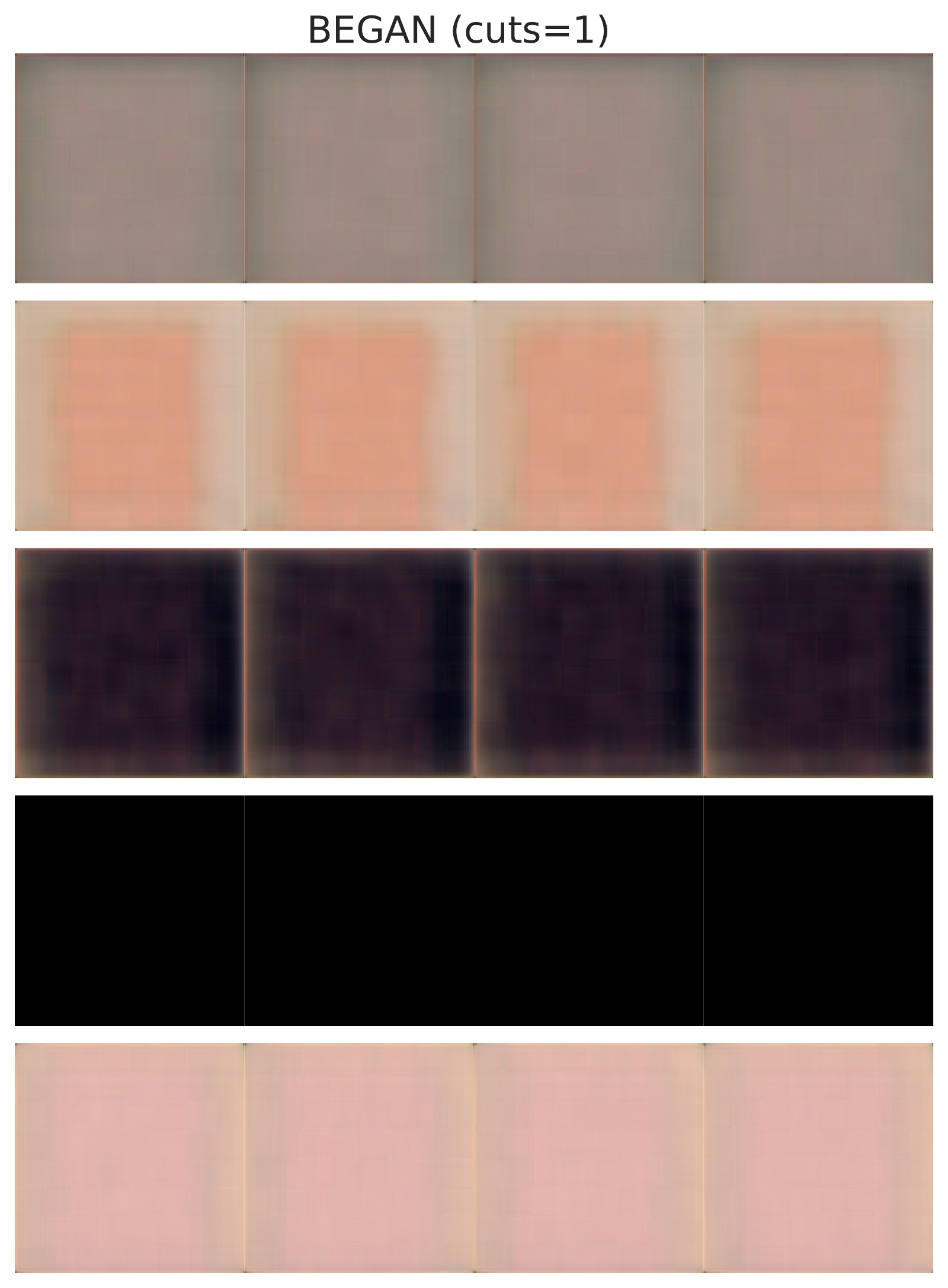}
\end{subfigure}
\caption{Models cut before training ($c=1$). Each row represents one hyperparameter setting, each column is a different random sample. Generated samples show failed training.}
\label{fig:cut-training}
\end{figure}

To confirm that layers cannot be cut at training time, we try training BEGAN and DCGAN with a variety of hyperparameter settings for at least 20 epochs \emph{after} cutting $c=1$ \layers{}. 
In Fig.~\ref{fig:cut-training}, each row represents a single hyperparameter setting and each column is an independent random sample.
All attempts result in mode collapse with poor quality output images.
Note the stark difference when compared to the samples from successful training in the first rows of Fig.~\ref{fig:generator-samples}.

\subsection{Choosing Number of Blocks to Cut}
\begin{figure}[H]
\centering
\includegraphics[width=\linewidth]{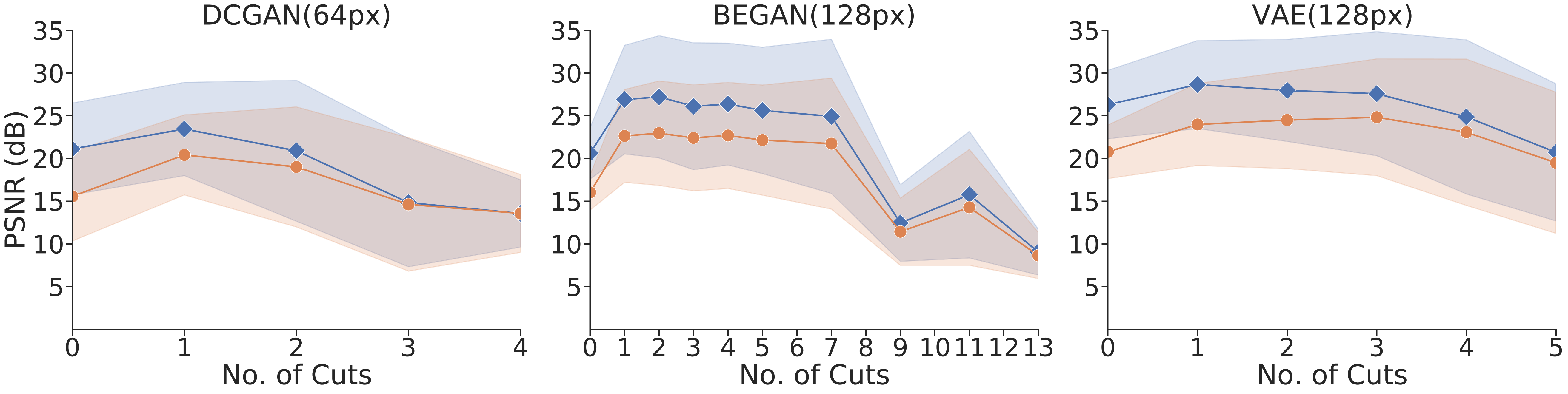}
\caption{Varying cut index $c$ of GS using DCGAN, BEGAN, and VAE architectures on CS.}
\label{fig:best-cuts}
\end{figure}

We find the cut index $c$ for each model by evaluating compressed sensing performance on validation images from CelebA and COCO. We use $100$ images and compute the average PSNR for each dataset. Fig.~\ref{fig:best-cuts} shows the average PSNR for these datasets when applying \ours{} to DCGAN, BEGAN, and VAE models. For all other experiments, we select the $c$ which achieves the highest PSNR on CelebA.

\subsection{Choice of Initialization}
\begin{figure}[h]
\begin{center}
\includegraphics[width=\columnwidth]{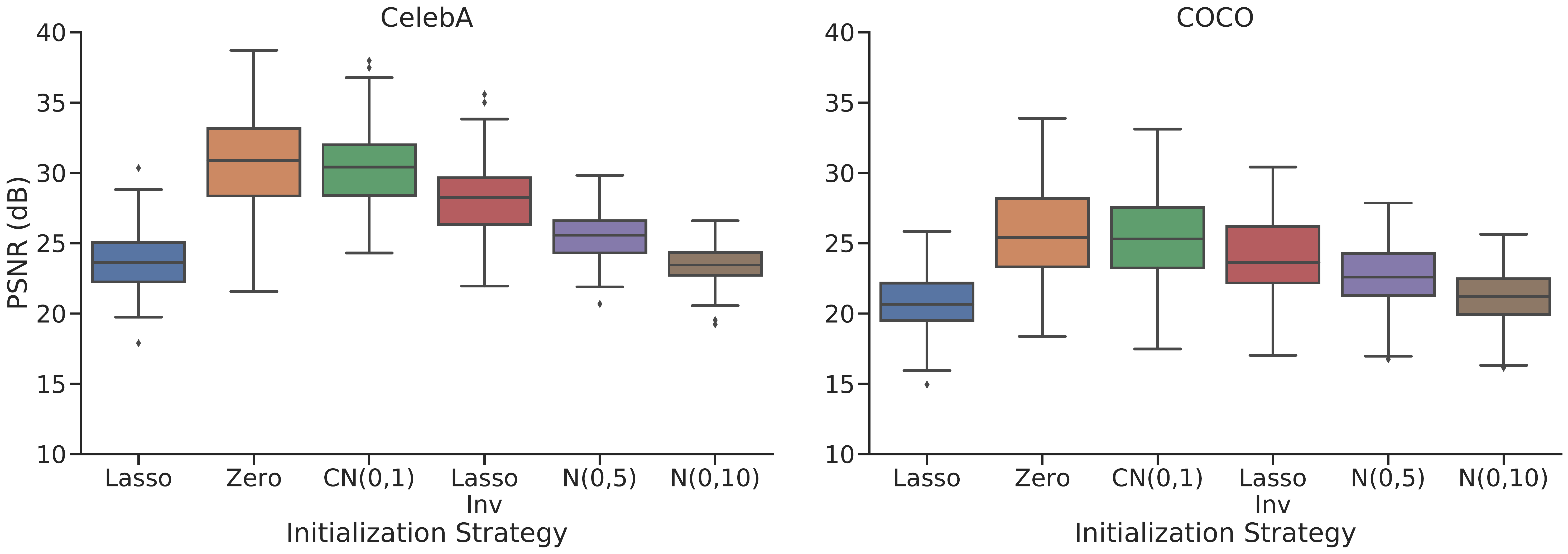}
\end{center}
\caption{Different initialization strategies for BEGAN with surgery on compressed sensing at $m/n \approx 0.16$ of $100$ images on CelebA and COCO. See text for label explanations.}
\label{fig:cs-other-init}
\end{figure}

We consider different initialization strategies in Fig.~\ref{fig:cs-other-init}. 
``Zero'' is the all-zeros vector and ``$CN(0,1)$'' denotes a standard normal distribution censored to the range $[-1, 1]$. ``LassoInv'' denotes using the solution to the Lasso-DCT optimization (the latent vector $z$) as initialization. $\mathcal{N}(0, 5)$, $\mathcal{N}(0, 10)$ are normal distributions with increasing variance. All initializations were run with the best of 3 random restarts. Performance differs according to the initialization strategy and we find that $CN(0,1)$ and ``Zero'' perform best, though ``Zero'' has slightly higher variance. We use $CN(0,1)$ for all experiments.

\end{document}